# A Review on Part-of-Speech Technologies

Onyenwe Ikechukwu, Onyedikachukwu Ikechukwu-Onyenwe, Onyedinma Ebele

Department of Computer Science, Nnamdi Azikiwe University, Awka, Nigeria
ie.onyenwe@unizik.edu.ng

## ABSTRACT

*Developing an automatic part-of-speech (POS) tagging for any new language is considered a necessary step for further computational linguistics methodology beyond tagging, like chunking and parsing, to be fully applied to the language. Many POS disambiguation technologies have been developed for this type of research and there are factors that influence the choice of choosing one. This could be either corpus-based or non-corpus-based. In this paper, we present a review of POS tagging technologies.*

## 1. INTRODUCTION

Part of speech (henceforth POS) tagging is the process of assigning a POS or other lexical class marker to each word in a language texts according to their respective POS label according to its definition and context [4] [14]. It is an important enabling task for NLP applications such as a preprocessing step to syntactic parsing, information extraction and retrieval, statistical machine translation, corpus linguistics, etc.

Automating POS tagging involves three main tasks: text segmentation into tokens, assignment of most probable tags to tokens (this create room for ambiguity), and determination of potential appropriateness of each probable tag to resolve ambiguity. There is valid contextual information used by all POS tagging technologies in their analysis for disambiguation, namely preceding (following) words, and their POS tags. This information is highly utilized in trying to resolve ambiguity. There is valid contextual information used by all POS tagging technologies in their analysis for disambiguation, namely preceding (following) words, andvtheir POS tags. This information is highly utilized in trying to resolve ambiguity. [2] justified the effectiveness of this approach in the following comment:

*"Corpus-based methods are often able to succeed while ignoring the true complexities of language, banking on the fact that complex linguistic phenomena can often be indirectly observed through simple epiphenomena"* [2]

Brill illustrated the above statement using "race" in examples 1–3, that one could assign a POS tag to words in the sentence considering the contextual information.

1. He will race/verb the car.

2. He will not race/verb the car.

3. When will the race/noun end?

The POS tags attached to the word "race" were derived considering the preceding (following) words and their associated tags based on the fact that words seen in the right side of a modal, in this case "will" is always a verb. This linguistic generalization is only exempted when there is a determiner to the right of the modal. And there is an assumption that determiner "a/the" precedes either a noun or adjective in English, but in this case, it is a noun. Therefore, restricting POS tagging to the use of this minimal information while ignoring the complexities found in most languages can be highly effective.

## 2. PART-OF-SPEECH TECHNOLOGY

Various POS technologies developed are either corpus-based or non-corpus-based. The above illustrations are examples of corpus-based method. [1] and [14] identified types of technologies developed and classified them on the following basis:

1. source of knowledge: here, linguistic generalization used in the model are derived from either the grammatical knowledge of a linguist or the text corpus.

2. Express as: the linguistic generalization derived are encoded as rules or probabilities (linguistic information are represented in numerical parameters).

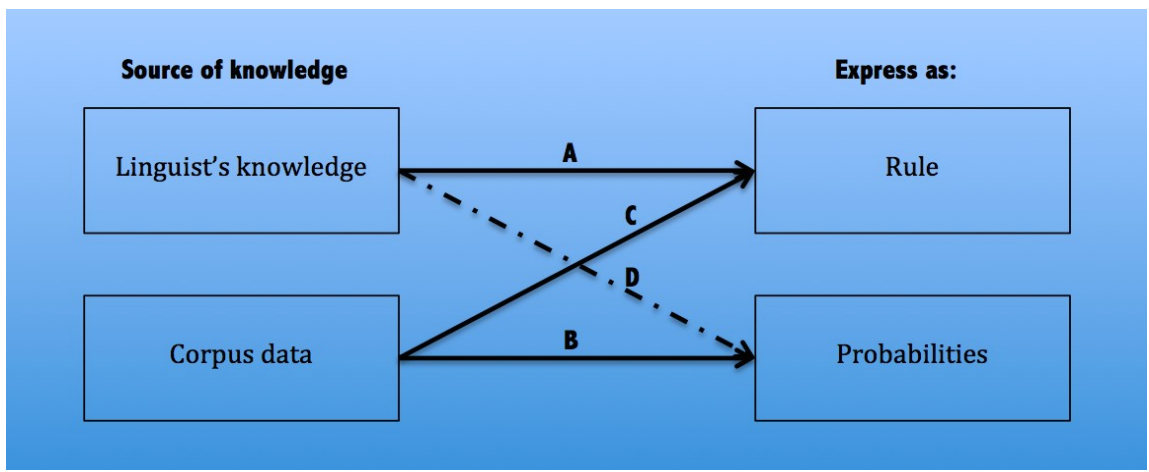

Logically, there are four possible POS tagging techniques in Figure 1, namely rule-based method (A), probabilistic-based method (B), corpus-based method (C), and unknown mode of operation (D). The illustration in the Figure 1 is closely in relation to four possible disambiguation methodologies in [1].

### 2.1. Rule-based method (A)

Rule-based (A) is where linguistic information is expressed as rules. This the earliest POS tagging technique developed, between 1960s and 1980s. It is a combination of lexicon and hand- crafted rules. Model is EngCG (English Constraint Grammar). In this approach, linguistic generalization forms the information called constraints. There are set of instructions in each rule that determine how operations are to be performed and contextual information that describe where to apply the rules. The operations tend to alter tags associated with ambiguous tagged terms so that one or more potential tags are eliminated from the list of consideration, thereby reducing ambiguity. Rule-based is more than using grammar as traditionally formulated, it involves syntactic analysis. The work of [15], "string analysis of sentence structure" is an example. It is described by Harris as "a process of syntactic analysis which is intermediate between the traditional constituent analysis and transformational analysis." Another example is constraint grammar (CG), a method implemented in the work of [10]. Karlsson defined CG as "a language-independent formalism for surface-oriented, morphology- based parsing of unrestricted text."

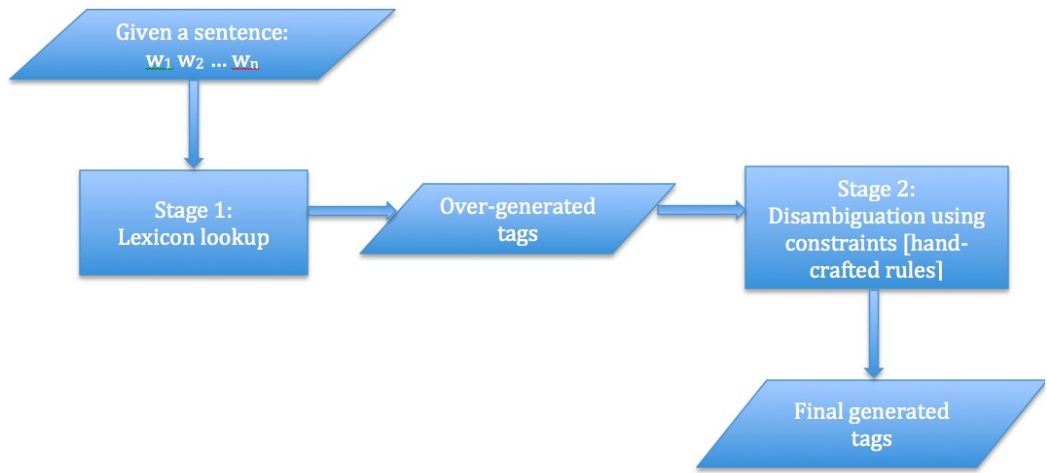

Figure 2: EngCG Architecture

CG has been implemented in English, Swedish, and Finnish. It is based on disambiguation of morphological readings and syntactic labels. The assignment of tags is directly through lexicon to syntax. Suggested alternative tags are discarded by the constraints as many as possible, the outcome aimed at getting fully disambiguated sentence having one tag for each word. Figure 2 shows the architecture of CG. CG rule-based approach has been successfully applied to most of European languages like English and French. The EngCG tagger using 1,100 constraints achieved a recall of at least 99.7% with 93-97% of words disambiguated [1]. Also, Greene and Rubin used a rule-based approach in the TAGGIT system [8], which aided in POS tagging the Brown corpus [7]. 77% of the corpus was disambiguated through the use of TAGGIT; the remaining ambiguated words in the corpus were manually corrected over a period of several years.

## 2.2. Probabilistic-based method (B)

This approach enables linguistic generalization derived from a corpus-data to be represented in numerical parameters. The frequencies at which different sequences of tags occur are derived from the corpus-data. The derived information is then used to verify "which of this possible tags is the most likely for w (observed word) in the vicinity $X$ ($X$ could be a sentence under observation) given a sequence of tags ($t_i...t_n$)". This statistical information gathered help in choosing the right tag of the ambiguously tagged word using any machine learning applicable design. Illustrating the above statement, a tagger after training on a gold corpus might learn through statistics that a tag for a subject pronoun is followed by the verb tag 70% of the time, adverb tag 29% of the time, and noun tag 1% of the time. In the future, if the tagger encounters a new instance, say, a word following a subject pronoun that was ambiguously tagged as either a noun or verb, it can use its statistical knowledge to decide that the verb tag is most likely to be correct [1]. Though in practice, this process would be incapable of handling long sequences of ambiguous words. Thus, the modern probabilistic process called Markov model. This model works by estimating the probability of a sequence of tags, given the transition probability of the different tags in the sequence $P(t_i|t_{i-1})$. Markov model uses transition probabilities and observation likelihood $P(w_i|t_i)$ in calculating the possible tag for the ambiguously tagged words. Both probabilities correspond to the prior and likelihood respectively.

Given annotated corpus, to calculate the maximum likelihood estimate of $P(t_i|t_{i-1})$, we compute the counts of $t_i$ given $t_{i-1}$ and divide the outcome by count $t_{i-1}$, that is:

$$P(t_i|t_{i-1}) = C(t_{i-1}, t_i) \div C(t_{i-1}) \qquad (1)$$

For example, determiner is very likely to follow adjectives and nouns in English (e.g. that/DT flight/NN and the/DT yellow/JJ hat/NN). Thus, we would expect P(NN/DT) and P(JJ/DT) to have higher scores compare to P(DT/JJ) and P(DT/NN) using equantion 1 [60]. Also, in the same way, we can compute maximum likelihood estimate of a word w under observation. So, we count how many times w occur in the annotated corpus given a tag t. The outcome is divided by number of t in the corpus, that is:

$$P(w_i|t_i) = C(t_i, w_i) \div C(t_i) \qquad (2)$$

[9] stated "... in POS tagging, while we recognize the words in the input, we do not recognize the POS tags. Thus, we cannot condition any probabilities on, say, a previous POS tag, because we cannot be completely sure exactly which tag applied to the previous word." Hence, the term hidden in Markov model because the transitions are hidden from view. Hidden Markov Model (HMM) handles both seen and hidden events. Observed events are the words that serve as an input while hidden events are the POS tags. The Markov model parameters useful in POS tagging are the states are the tags, the symbols that they output (their associated words), and the transitions from one state to another (makeup the sequence of tags allocated to a sentence). Tag transition probabilities answer the question "how likely are we to expect a tag given the previous tag?" and lexical likelihood finds how likely "the observed word" given a POS tag. Final estimation is computed by calculating the maximum likelihood for all the probabilities which can be obtained from the corpus counts. The most likely tag in Markov model can be found through computation or selecting the most appropriate path without calculating the probability of each tag. Using the example in [9], the former works by focusing on the observed word and the tag sequence. For instance, the word "race" can be a noun or a verb in the followings:

1. Secretariat/NNP is/BEZ expected/VBN to/TO race/VB tomorrow/NR.

2. People/NNS continue/VB to/TO inquire/VB the/AT reason/NN for/IN the/AT race/NN for/IN outer/JJ space/NN.

A graph view of 1 and 2 above is gven in Figure 3

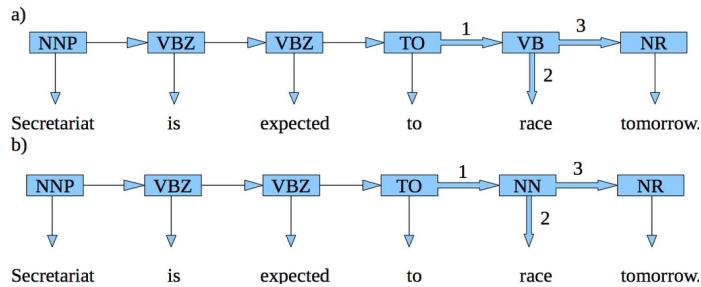

Figure 3: Graph showing three probabilities (arcs 1, 2 and 3) where sentences 1 and 2 above differ (Source: [9])

Arcs 1, 2 and 3 in Figure 3 are the probabilities that HMM will use to resolve ambiguously POS tagged words. Like in the above case *(race/NN, race/VB)*. Arcs 1 and 2 are the transition probability $P(t_i|t_{i-1})$ and lexical likelihood $P(w_i|t_i)$, while 3 is the probability of the tag sequence in case of NR. So, the transition probabilities of *P(VB|TO)* and *P(NN|TO)*, assume that verbs are about 500 times as likely as nouns to occur after TO, using equation 1 is as follows, *P(VB|TO) = 0.83* and *P(NN|TO) = 0.00047*. The lexical likelihood of "race" given a tag (i.e. the probability of tags *VB* and *NN* producing the word race) using equation 2, P(race|NN) = 0.00057 and P(race|VB) = 0.00012. And finally, the probability of the tag sequence (arc 3 in figure 3.4), P(NR|VB) = 0.0027 and P(NR|NN) = 0.0012. To get the final probability, multiply all the probabilities to get

- P(VB|TO)P(NR|VB)P(race|VB) = 0.00000027

- P(NN|TO)P(NR|NN)P(race|NN) = 0.00000000032

With this statistic, HMM tagger will correctly tag race as a VB. Markov model has been widely used in assuming that a token depends probabilistically on its part-of-speech category and this in turn depends solely on the categories of the following and preceding one or two words [5]. Markov model are also called n − gram taggers because of the transition probabilities, which might occur over n − size of words, taggers based on. There is unigram tagger where n = 1, bigram taggers where n = 2 and trigram taggers where n = 3. These taggers undergo a training process to learn statistical behaviour of the tags associated with words in the corpus data through an examination of the training data (manually tagged corpus). These statistical learned behaviour of the tagger are used in disambiguation techniques, in Markov model. Viterbi algorithm allows an appropriate tag to be selected through choosing the most probable path without calculating the probability of each path through the ambiguously tagged sequences of words. It is the most decoding algorithm used in HMMs, mostly relevant in speech and language processing. According to [12], there are three reasons of choosing Viterbi over maximum likelihood. Firstly, it is easier to implement. Secondly, it cannot produce in its output sequences of tags which are not possible in the model, which is theoretically possible in maximum likelihood tagging. Thirdly, it gives the best analysis of the sentence as a whole.

Stochastic process of POS tagging advantage over rule-based is that the linguist does not write an effective set of rules to produce an effective system. According to [2], the stochastic techniques advantage over the traditional rule-based design comes from the ease that very little hand-crafted knowledge need be built into the system. Also, stochastic process is generally applicable. Probabilistic or stochastic process became popular in 1980s. Statistical methods have been used in the works of [6] (VOLSUNGA), [12], [5], [3] (TnT), etc.

## 2.3. Corpus-based Rule Method (C)

In this approach, linguistic generalization (rules) are generated from the corpus, hence the name corpus-based rule. Corpus-based rule ignores the true complexities of language in its process with the fact that challenging linguistic phenomena can be indirectly recognized through simple epiphenomena [2]. But this does not mean that the rule-based approach discussed earlier did not use corpus data to derive some grammatical rules. [11] pointed out that some linguistic generalization used in CGC system were based on empirical observation of the corpus data. However, both approaches are distinguished in their implementations. Corpus-based rule method allow computer to formulate and evaluate the rules. That is, the rules created are automatic and linguistic readable. The linguist information about grammar are excluded from this process. An example of this approach is a widely used Brill POS tagger based on transformation- based error-driven learning.

## 2.4. Unknown mode of operation (D)

Dashed line in Figure 1 above indicates unknown mode of operation. According to [1], humans are unskilled in estimating frequencies of words and other linguistic phenomena. Therefore, modelling linguist knowledge of the corpus data in probabilistic form will create a perverse POS tagging techniques. Since humans will usually get words frequencies wrong compared to rules, which are more likely to get it right.

## 3. CONCLUSION

There are different approaches to the problem of assigning a part of speech (POS) tag to each word of a natural language sentence. These different approaches fall into two distinctive groups: corpus-based and non-corpus-based. Here, we present an elaborate review of the existing technologies in the area of POS tagging with a focus on these distinctive groups.